\documentclass[10pt,twocolumn,letterpaper]{article}

\usepackage[usenames,dvipsnames,table]{xcolor}
\usepackage{iccv}
\usepackage{times}
\usepackage{epsfig}
\usepackage{graphicx}
\usepackage{amsmath}
\usepackage{amssymb}
\usepackage{tabularx} 
\usepackage{url}
\usepackage{array}
\usepackage{booktabs}
\usepackage{subcaption}
\usepackage{enumitem}
\usepackage{pifont}% http://ctan.org/pkg/pifont
\newcommand{\cmark}{\ding{51}}%
\newcommand{\xmark}{\ding{55}}%
\usepackage{booktabs,amsfonts,dcolumn} % HY: delete subcaption

\usepackage[space, compress, sort]{cite}
\newcommand\blfootnote[1]{%
\begingroup
\renewcommand\thefootnote{}\footnote{#1}%
\addtocounter{footnote}{-1}%
\endgroup
}

\usepackage[pagebackref=true,breaklinks=true,letterpaper=true,colorlinks,bookmarks=false ]{hyperref}
\hypersetup{citecolor=[RGB]{119,185,0}}

\iccvfinalcopy % *** Uncomment this line for the final submission

\def\iccvPaperID{2827} % *** Enter the ICCV Paper ID here

\ificcvfinal\fi

\begin{document}

%%%%%%%%% TITLE
\title{FB-BEV: BEV Representation from Forward-Backward\\ View Transformations}

\author{
Zhiqi Li$^{1,2*}$~
Zhiding Yu$^2$~
Wenhai Wang$^{3}$~
Anima Anandkumar$^{2,4}$~
Tong Lu$^{1}$~
Jose M. Alvarez$^2$
\\ [0.15cm]
$^1$National Key Lab for Novel Software Technology, Nanjing University~~$^2$NVIDIA \\
$^3$The Chinese University of Hong Kong~~$^4$Caltech
}

\ificcvfinal\thispagestyle{empty}\fi

\twocolumn[{%
	\renewcommand\twocolumn[1][]{#1}%
	\maketitle
	\centering
	\vspace{-0.5cm}
	\begin{minipage}{0.33\linewidth}
		\includegraphics[width=\linewidth]{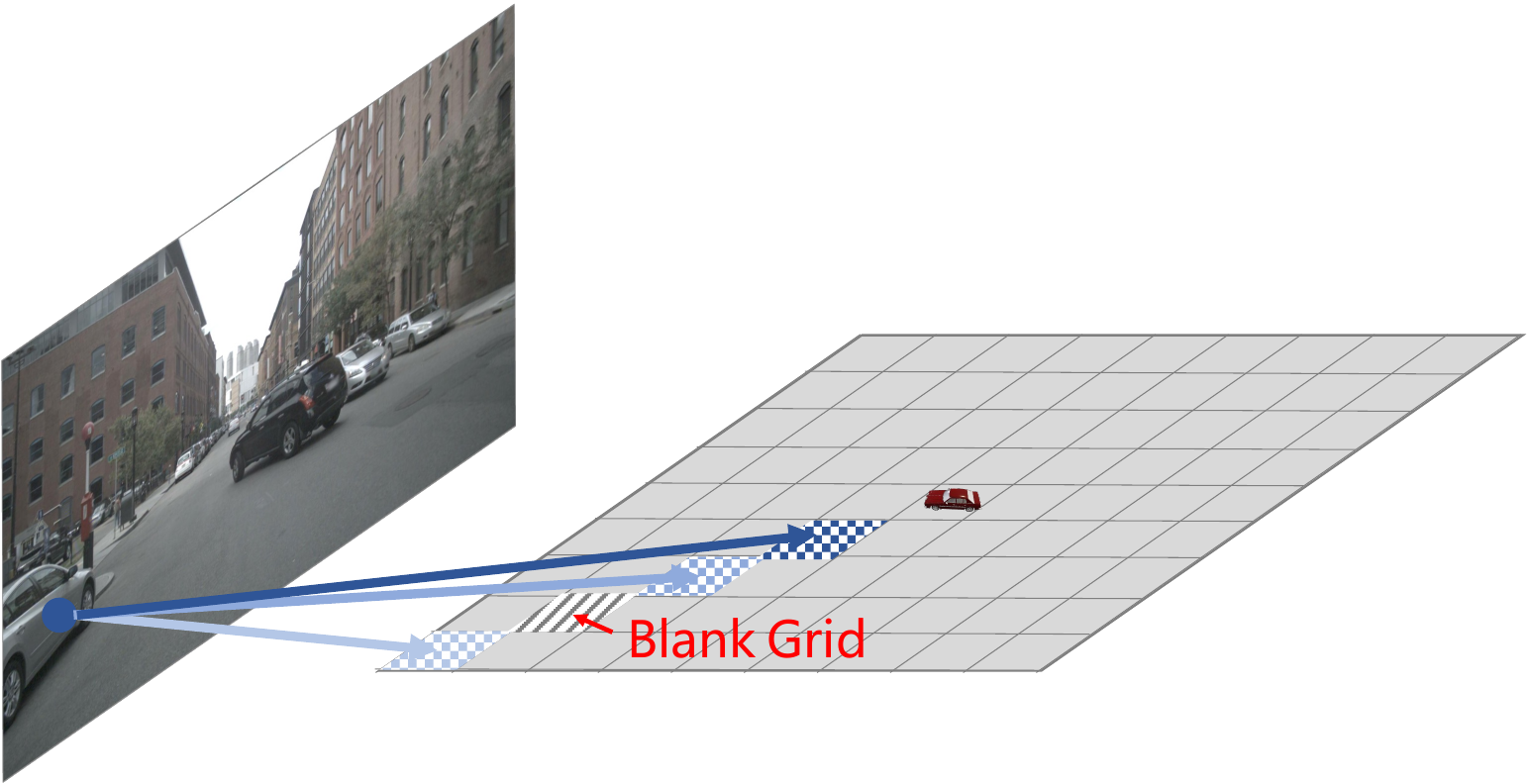}
		\label{fig1:sub1}
	\end{minipage}
	\begin{minipage}{0.33\linewidth}
		\includegraphics[width=\linewidth]{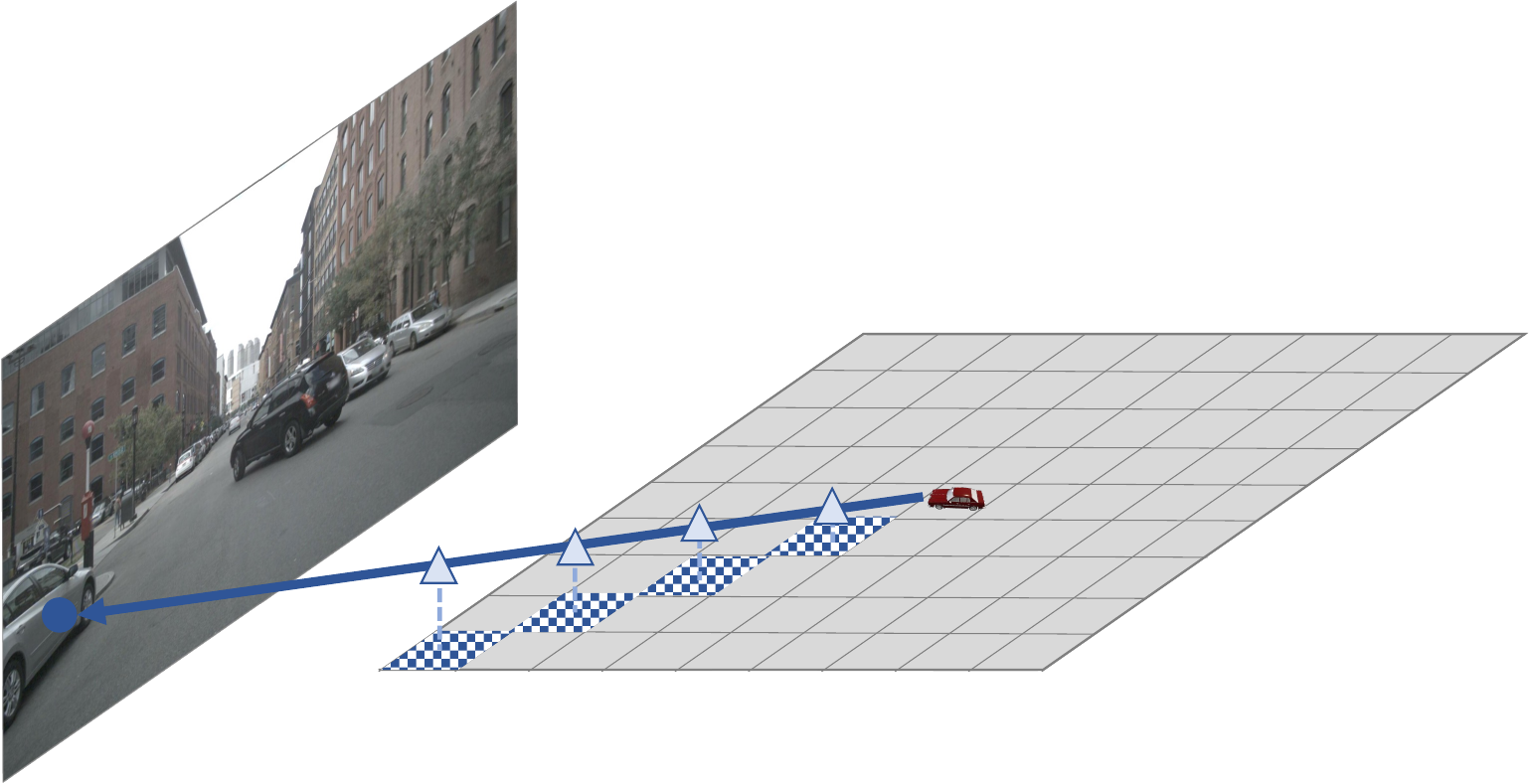}
		\label{fig1:sub2}
	\end{minipage}
	\begin{minipage}{0.33\linewidth}
		\includegraphics[width=\linewidth]{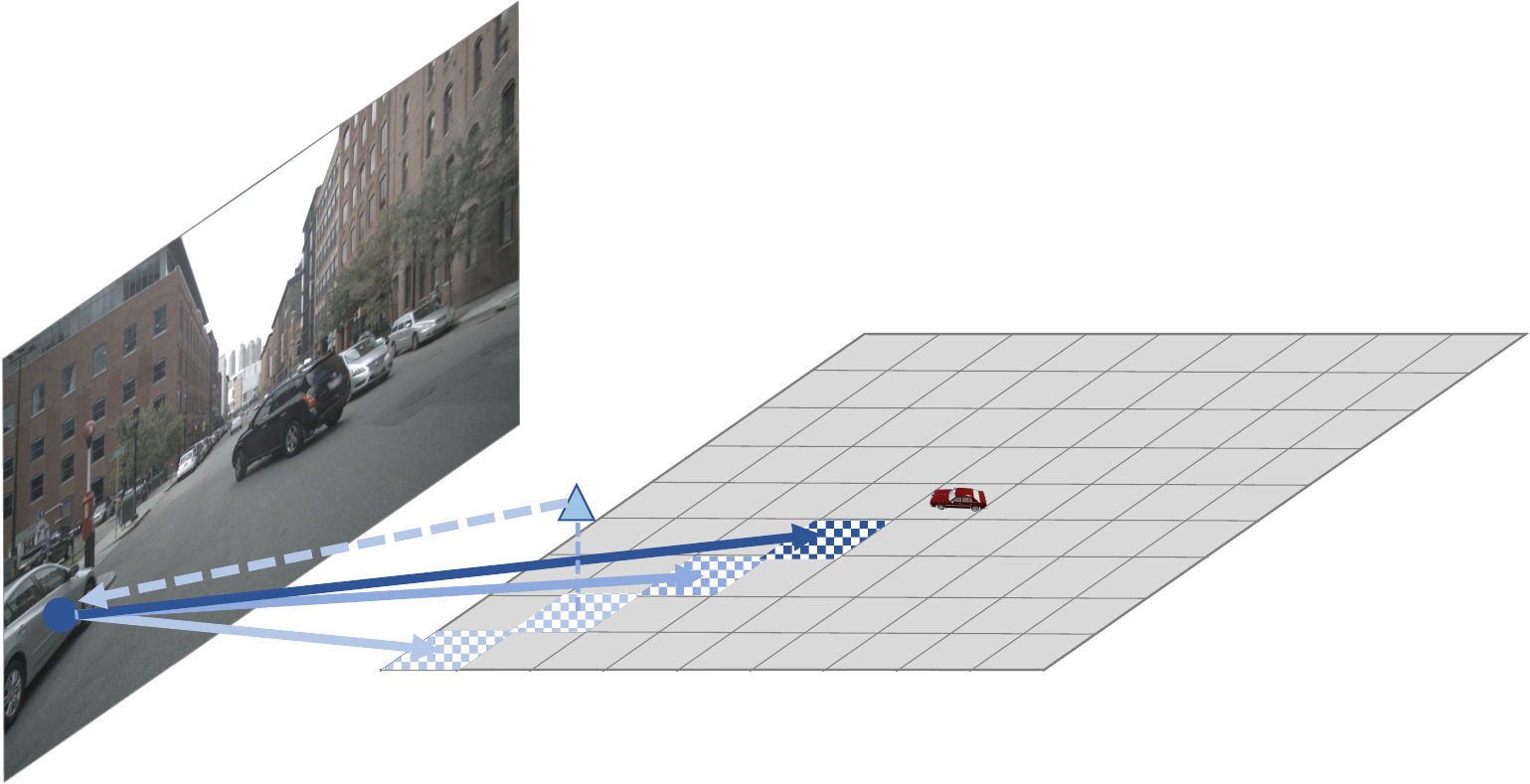}
		\label{fig1:sub3}
	\end{minipage}
	\vspace{-0.5cm}
	\captionof{figure}{
		\textbf{Left}: Forward projection lifts the image features from 2D to BEV space and weight them based on depth (coded in different colors). However, forward projection tends to generate sparse BEV projection.
		\textbf{Middle}: Backward projection first defines the voxel locations in the 3D space and then projects these points onto the 2D image planes. Dense BEV features can be generated but the points on the projection ray fetch the same features without distinction.
		\textbf{Right}: Forward-backward projection proposed in this work. We use backward projection to refine the necessary BEV grids with reduced sparsity. We further introduce depth consistency into backward projection and assign each projection a different weight (dashed line).
	}
	\label{fig:moti}
	\vspace{0.55cm}
}]

\begin{abstract}
View Transformation Module (VTM), where transformations happen between multi-view image features and Bird-Eye-View (BEV) representation, is a crucial step in camera-based BEV perception systems. Currently, the two most prominent VTM paradigms are forward projection and backward projection. Forward projection, represented by Lift-Splat-Shoot, leads to sparsely projected BEV features without post-processing. Backward projection, with BEVFormer being an example, tends to generate false-positive BEV features from incorrect projections due to the lack of utilization on depth.
To address the above limitations, we propose a novel forward-backward view transformation module. Our approach compensates for the deficiencies in both existing methods, allowing them to enhance each other to obtain higher quality BEV representations mutually. We instantiate the proposed module with FB-BEV, which achieves a new state-of-the-art result of 62.4\% NDS on the nuScenes test set. Code and models are available at \url{https://github.com/NVlabs/FB-BEV}.\blfootnote{* Work done during an internship at NVIDIA.}
\end{abstract}

%%%%%%%%% BODY TEXT

BEV-based 3D detection models have gained popularity due to their unified and comprehensive representation abilities for multi-camera inputs, enhancing the performance of both vision-only and multi-modality perception models for autonomous driving\cite{philion2020lift, wang2022detr3d,li2022BEVFormer, liu2022bevfusion,liang2022bevfusion,bai2022transfusion, li2022delving, rukhovich2022imvoxelnet, huang2021bevdet}. A typical BEV-based detection model comprises an image backbone, a View Transformation Module (VTM), and a detection head. The VTMs primarily function to project multi-view camera features onto the BEV plane. There are two main categories of existing mainstream VTMs based on the projection methods used: forward projection and backward projection.

\section{Introduction}
\begin{figure}[t]
	\centering
	\begin{minipage}{0.5\linewidth}
		\centering
		\includegraphics[width=0.92\linewidth]{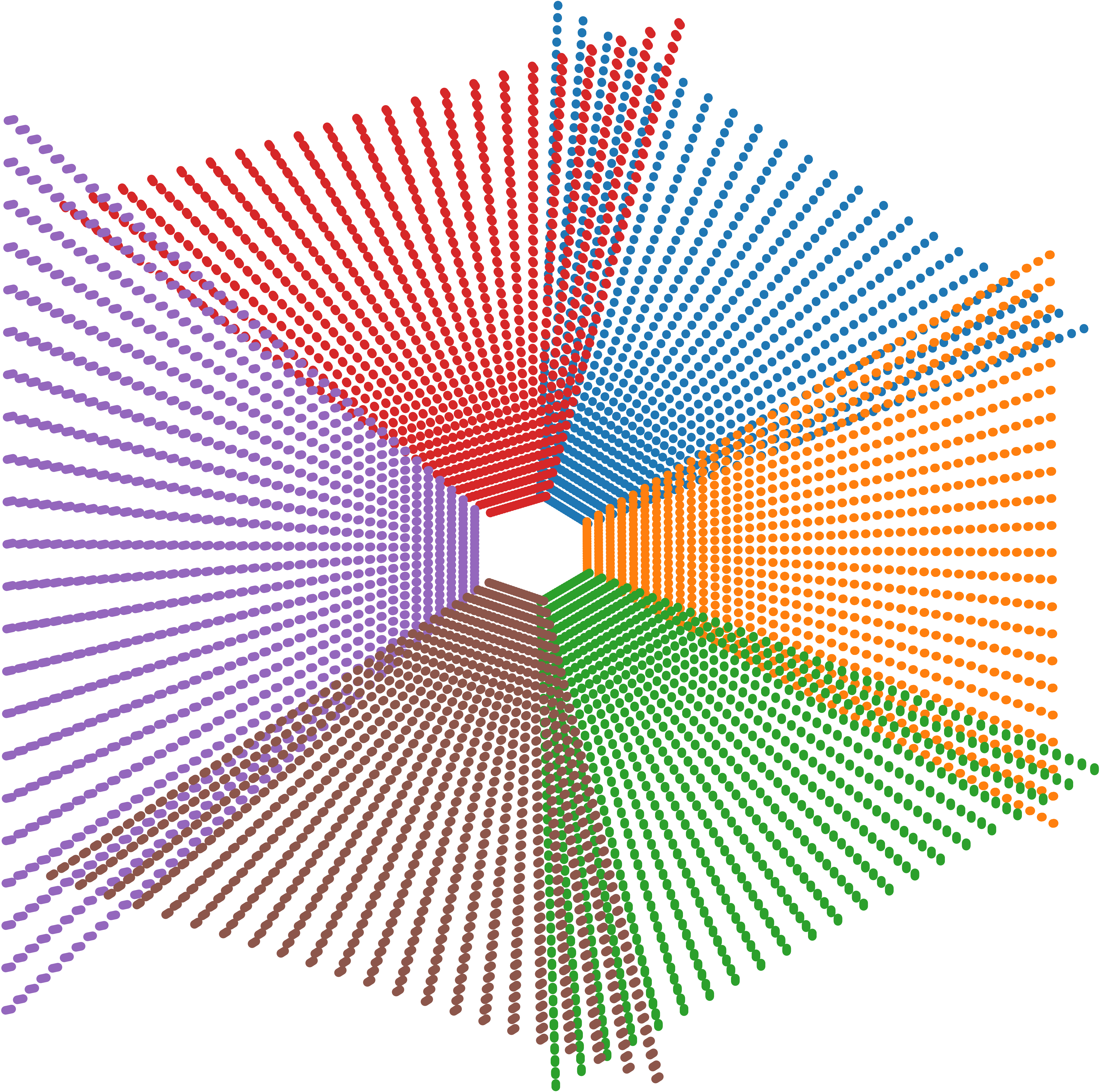}
		\subcaption{}
		\label{fig2:sub1}
	\end{minipage}%
	\begin{minipage}{0.5\linewidth}
		\centering
		\includegraphics[width=0.92\linewidth]{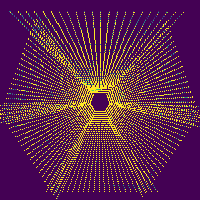}
		\subcaption{}
		\label{fig2:sub2}
	\end{minipage}
	\vspace{-1mm}
	\caption{(a) Projection points on BEV plane with forward projection on nuScenes dataset, and different colors represent different cameras. (b) BEV feature map of LSS~\cite{philion2020lift} with a shape of 200$\!\times\!$200. We can observe that forward projection has an extremely low utilization rate for BEV space.}
	\label{fig:lss}
	\vspace{-0.1in}
\end{figure}

\paragraph{Forward projection.} 
The most intuitive method for projecting camera features onto the BEV plane involves estimating the depth value of each pixel in the image and using the camera calibration parameters to determine the corresponding position of each pixel in 3D space~\cite{wang2019pseudo}, as shown in Figure~\ref{fig:moti} (left).  We refer to this process as forward projection, where the 2D pixels take the initiative in projection and the 3D space passively accepts features from the images. The accuracy of the predicted depth for each pixel is critical to achieving high-quality BEV features. However, accurately estimating the depth value of each pixel is challenging~\cite{park2021pseudo}.
To address this challenge, Lift-Splat-Shoot (LSS) pioneered the use of depth distribution to model the uncertainty of each pixel's depth~\cite{philion2020lift}. 
One limitation of LSS is that it generates discrete and sparse BEV representation~\cite{philion2020lift, zhou2022matrixvt}.
As shown in Figure~\ref{fig:lss}, the density of BEV features decreases with distance. When using the default settings of LSS on the nuScenes dataset, only 50\% of the grids can receive valid image features through projection.

\paragraph{Backward projection.}
The motivation behind backward projection is opposite to that of forward projection. For the backward projection paradigm, the points in 3D space take the initiative~\cite{rukhovich2022imvoxelnet, Roddick2019OrthographicFT, li2022BEVFormer,wang2022detr3d}. For instance, BEVFormer sets the coordinates of the 3D space to be filled in advance and then projects these 3D points back onto the 2D image~\cite{li2022BEVFormer}, as shown in Figure~\ref{fig:moti} (middle). As a result, each predefined 3D space position can obtain its corresponding image features. The BEV representation obtained by this method is denser than that of LSS, with each BEV grid filled with the corresponding image features. 

\begin{figure}[t]
	\centering
	\begin{minipage}{0.5\linewidth}
		\centering
		\includegraphics[width=0.95\linewidth]{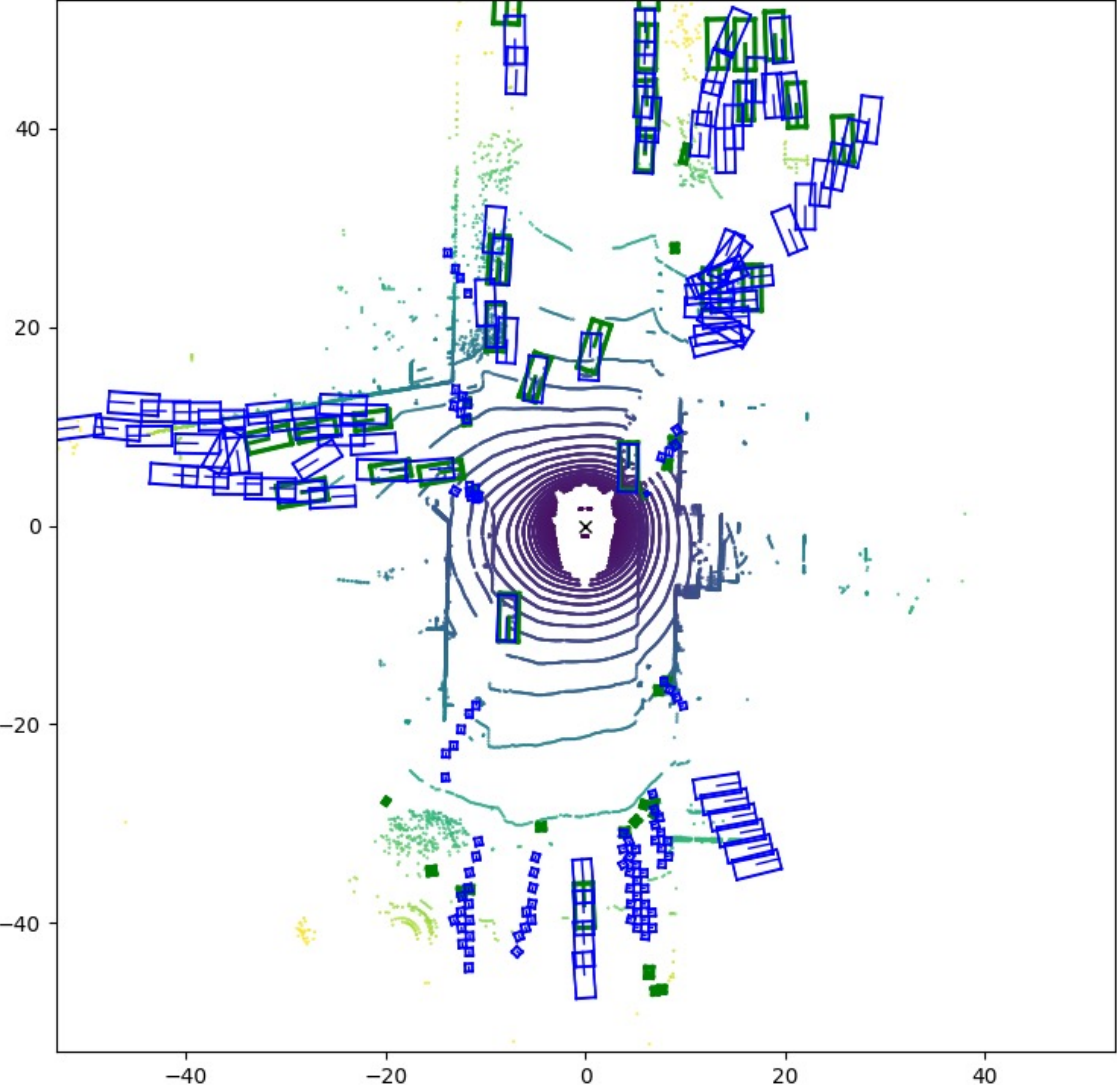}
		\subcaption{}
		\label{fig3:sub1}
	\end{minipage}%
	\begin{minipage}{0.5\linewidth}
		\centering
		\includegraphics[width=0.95\linewidth]{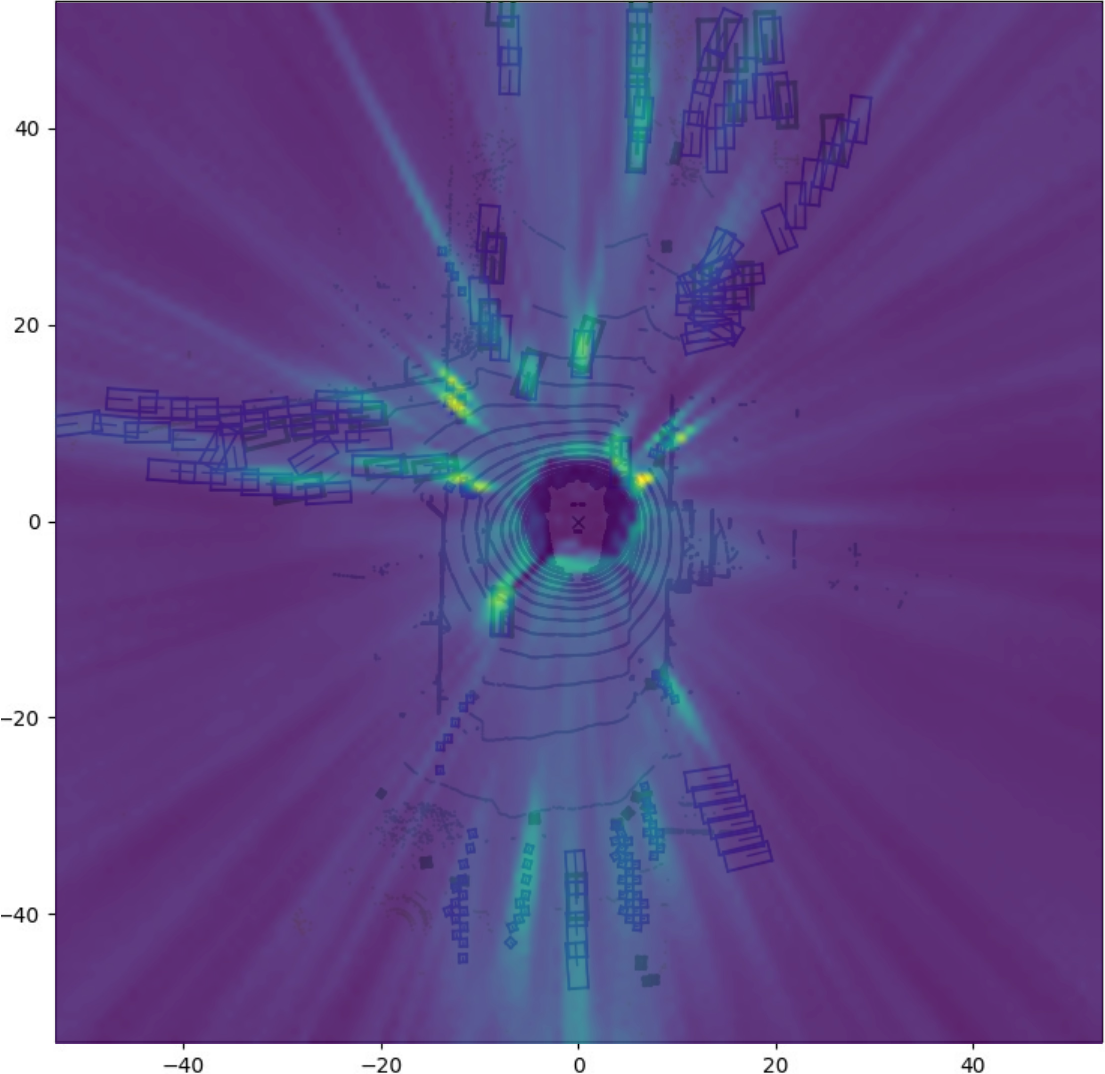}
		\subcaption{}
		\label{fig3:sub2}
	\end{minipage}
	\vspace{-1mm}
	\caption{(a) Detection of BEVFormer. (b) The corresponding BEV features of BEVFormer. Since BEVFormer cannot use depth for distinction, the features of each object on the BEV tend to be ray-shaped. The model thus predicts multiple boxes for one object along the longitudinal direction.}
	\label{fig:3d_to_2d_vis}
	\vspace{-0.1in}
\end{figure}

The drawbacks of backward projection are also apparent, as shown in Figure~\ref{fig:3d_to_2d_vis}. Although yielding a denser BEV representation, it comes at the cost of establishing numerous false correspondences between 3D and 2D space due to occlusion and depth mismatch~\cite{hu2020you}. The absence of depth information during the projection process is the main cause.  Without depth as a reference, each 3D coordinate on the ray is equally related to the same 2D coordinate, equivalent to having a uniform depth distribution for this pixel in forward projection. As a result, the distance prediction of the objects along the longitudinal direction become ambiguous. Backward projection thus tends to be inferior to forward projection in depth utilization. Recently, the advantage of forward projection has been further highlighted since more accurate depth distribution obtained from depth supervision is shown to improve 3D perception~\cite{li2022bevdepth,li2022bevstereo}.

Considering the pros and cons discussed above, we propose forward-backward view transformation to address the limitations of existing VTMs, as shown in Figure~\ref{fig:moti} (right). To address the issue of sparse BEV representations in forward projection, we leverage backward projection to refine the sparse region from forward projection. Meanwhile, backward projection is prone to false-positive features due to the lack of depth guidance. We thus propose a depth-aware backward projection design to suppress false-positive features by measuring the quality of each projection relationship through depth consistency. The depth consistency is determined by the distance of depth distributions between a 3D point and its corresponding 2D projection point. Using this depth-aware method, unmatched projections are given lower weights, which reduces the interference caused by false-positive BEV features.
In addition, for the objection detection task, we only care about foreground objects,  so we densify only the foreground regions of the BEV plane while using backward projection. This not only reduces the computational burden but also avoids the introduction of false-positive features in the background areas. With the sparse regions refined for forward projection and false-positives features reduced for backward projection, our forward-backward projection not only solves the defects of existing projection methods but also realizes the effective ensemble of existing projection methods. Our contributions can be summarized as follows:

\begin{itemize}[leftmargin=1.3em]
\item We propose a forward-backward projection strategy that generates dense BEV features with strong representation ability through bidirectional projection. Our approach addresses the limitations of existing projection methods, which result in either sparse BEV features or false-positive features caused by inaccurate projection.

\item To address the pitfalls of existing forward projection methods for producing sparse BEV representations, we employ backward projection to refine the blank grid that not be activated by forward projection. This makes the model more suitable for large-scale BEVs.

\item We propose a novel depth-aware backward projection method that overcomes the limitations of existing methods in effectively utilizing depth information. Our approach integrates depth consistency into the projection process to establish a more accurate mapping relationship between the 3D and 2D spaces.

\item Our FB-BEV model has been extensively evaluated on the nuScenes dataset. The results demonstrate that it outperforms other methods for camera-based 3D object detection and achieves the state-of-the-art 62.4\% NDS on the nuScenes \textit{test} set.
\end{itemize}

\begin{figure*}[t]
\begin{center}
\includegraphics[width=0.9\linewidth]{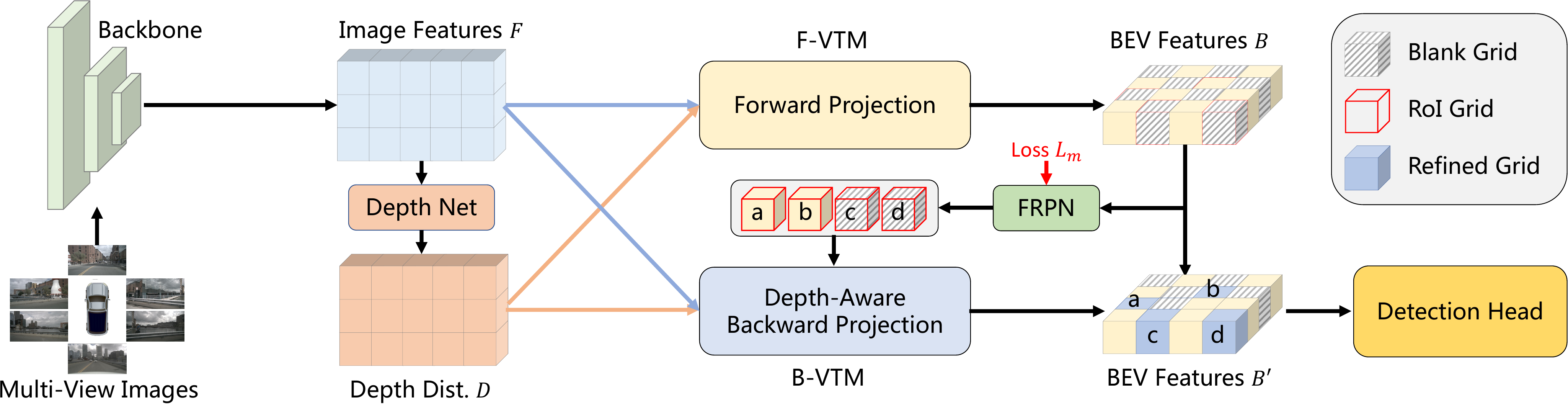}
\end{center}
 \vspace{-0.1in}
   \caption{Overview of FB-BEV. We first extract multi-view features from the 2D  backbone and generate the depth distribution using a depth network. We then employ a forward projection module to generate the BEV features $B$. Since BEV features $B$ contain blank grids, FRPN generates a foreground mask and feeds foreground region of interest (RoI) grids to the next depth-aware backward projection module (Grids $\{a,b,c,d\}$ in the figure).  Our depth-aware backward projection module uses RoI grids as BEV queries and refines these queries by projection them back onto images with a depth consistency mechanism.  Finally, we obtain the BEV features $B'$ by adding the refined grids and BEV features $B$. 
   }
\label{fig:arch}
  
\end{figure*}

\section{Related work}

We introduce related BEV perception works according to the VTMs they use.

\subsection{Forward Projection Methods}
The Lift-Splat-Shoot (LSS)~\cite{philion2020lift} method is the archetypal technique of this category. LSS utilizes a depth distribution to model depth uncertainty and project multi-view features into the same Bird's Eye View (BEV) space. Subsequent methods have largely adhered to this paradigm. For instance, BEVDet~\cite{huang2021bevdet} applies this forward projection approach to the field of multi-view 3D detection. CaDDN~\cite{reading2021categorical} and 
BEVDepth~\cite{li2022bevdepth} proposes the use of LiDAR point clouds to generate depth ground truth for supervising the depth prediction module. BEVDepth demonstrates that an accurate depth prediction module can significantly enhance model performance. Similarly, BEVstereo~\cite{li2022bevstereo} further underscores the importance of precise depth estimation to model performance. Furthermore, BEVFusion~\cite{liu2022bevfusion} extends this paradigm to the multi-modality perception domain and improves the projection efficiency of the LSS paradigm.
The most notable disadvantage of VTM in LSS is low efficiency. Subsequent research has made significant progress in improving efficiency through engineering implementation~\cite{huang2021bevdet, li2022bevdepth,liu2022bevfusion}. In response to the sparseness of BEV features, MatrixVT~\cite{zhou2022matrixvt} mainly focuses on improving the calculation efficiency in the process of BEV generation, rather than densely stressing BEV features.

\subsection{Backward Projection Methods}
OFT~\cite{Roddick2019OrthographicFT} is among the first methods to adapt the backward projection paradigm. 
This paradigm does not involve complex accumulation in 3D space~\cite{philion2020lift}, which is the least efficient step in forward projection. 
Subsequent works such as ImVoxelNet~\cite{rukhovich2022imvoxelnet} and M$^2$BEV~\cite{xie2022m2bev} extend this paradigm from monocular to multi-view perception where 3D space is divided into voxels. 
DETR3D~\cite{wang2022detr3d} does not introduce dense BEV features, but performs end-to-end learning of object queries in 3D and projects object centers back to image space.
BEVFormer~\cite{li2022BEVFormer} aggregates features at different heights on the BEV space without introducing voxelized representation, therefore reducing the resource consumption. BEVFormer also introduces deformable sampling points and temporal features, promoting further development of camera-based perception. For the perception heads, BEVFormer adopts Deformable DETR~\cite{zhu2020deformable} and Panoptic SegFormer~\cite{li2022panoptic}.
BEVFormerV2~\cite{yang2022BEVFormer} further exploits the potential of backward projection by adapting the modern image backbone via perspective supervision. PolarFormer~\cite{jiang2022polarformer} and PolarDETR~\cite{chen2022polar} adopt polar coordinates rather than Cartesian coordinates to conduct the projection process. Methods~\cite{lin2022sparse4d, qin2022uniformer,luo2022detr4d} project 3D anchors onto 4D features rather than 3D features.  PersFormer~\cite{chen2022persformer} uses Inverse Perspective Mapping (IPM) to guide the projection point on the image space.
However, existing methods seldom consider introducing depth in the projection process or even consider getting rid of the dependence on depth as an advantage~\cite{li2022BEVFormer, wang2022detr3d, xie2022m2bev}. We argue that it increases ambiguity in the projection process without depth to measure the quality of the projection.

In addition to different view projection paradigms, researchers have also explored using longer temporal information to enhance the spatial perception capacity.~\cite{park2022time,wang2023exploring, lin2023sparse4d,wang2023memoryandanticipation}.

\subsection{Projection-Free Methods}
In addition to the above two paradigms, some  methods can generate BEV representations without relying on projections.  PETR~\cite{liu2022petr} and PETRv2~\cite{liu2022petrv2} implicitly learn the view transformation through global attention and use camera parameters to encode position features.
CFT~\cite{jiang2022multi} uses view-aware attention to adaptively learn the BEV features required for each view, and even get rid of the dependence on camera calibration parameters.
BEVSegFormer~\cite{peng2023bevsegformer}  automatically learns the correspondence between 3D and 2D space without relying on the projection process. 

\section{Method}
To address the limitations of existing view transformation modules, we propose a novel Forward-Backward View Transformation method named FB-BEV. FB-BEV employs a two-pronged approach. Firstly, a VTM based on forward projection will generate an initial sparse BEV representation.  To obtain a denser BEV representation while minimizing the computational burden, a foreground region proposal network is employed to select the foreground BEV grids. Subsequently, another VTM utilizes these foreground grids as BEV queries and refines them by projecting them back onto the images with a depth-aware mechanism.

\subsection{Overall Architecture}
As illustrated in Figure~\ref{fig:arch}, FB-BEV mainly consists of three key modules: a view transformation module with forward projection denoted as F-VTM, a foreground region proposal network denoted as FRPN, and a view transformation module with depth-aware backward projection denoted as B-VTM. In addition, we have a depth net to predict the depth distributions, and the distributions will be utilized in both VTMs. F-VTM generates a complete BEV representation from the multi-view features by projecting each pixel feature into the 3D space based on the corresponding depth distribution. FRPN is a lightweight binarized mask predictor used to select the regions where the foreground object is located. 
B-VTM is only responsible for optimizing BEV grids located in the foreground region generated by FRPN.

During inference, we feed multi-view RGB images to the image backbone network and obtain the image features $F\!=\!\{F_i\}_{i=1}^{N_c}$, where $F_i$ is the view features of $i$-th camera view and $N_c$ is the total number of cameras. 
Then we obtain the depth distributions $D = \{D_i\}_{i=1}^{N_c}$ by feeding image feature $F$ into 
depth net.
Taking the view feature $F$ and depth distribution $D$ as input, the F-VTM will generate a BEV representation $B\in \mathbb{R}^{C\times H \times W}$, where $C$ is the channel dimension, and $H\times W$ is the spatial shape of BEV. 
FRPN takes BEV features $B$ as input and predicts a binary mask $M\in\mathbb{R}^{H\times W}$ to detect foreground regions. Only foreground grid $B[\text{sigmoid}(M)>t_f]$ will be fed in B-VTM to be further refined, where $t_f$ is the foreground threshold.  
The final BEV features $B'\in \mathbb{R}^{C\times H \times W}$ are obtained by adding the refined BEV features from B-VTM to $B$ back. Finally, we perform 3D detection task based on the BEV features $B'$.

\subsection{Forward Projection}
Our forward projection module F-VTM follows the paradigm of LSS~\cite{philion2020lift}.
Lift and Splat are two fundamental steps in modern forward projection techniques used for view transformation. The Lift step projects each pixel in the 2D image onto the 3D voxel space based on its corresponding depth distribution. The Splat step aggregates the feature values of pixels within each voxel by sum pooling.
For specific implementation, our F-VTM is based on BEVDet~\cite{huang2021bevdet,huang2022bevdet4d} and BEVDepth~\cite{li2022bevdepth}, which represent the current state-of-the-art design of forward projection. We denote the BEV features from F-VTM as $B$.

\subsection{Foreground Region Proposal Network}
The BEV features obtained from F-VTM are sparse, and there are BEV grids that are not activated and thus contain blank information.
To obtain a stronger BEV representation, we expect to fill in these blank BEV grids. However, for 3D object detection, our interest lies only in the limited foreground objects that occupy a relatively small fraction of the BEV features. To locate these foreground objects within the BEV features, we utilize a simple segmentation network to generate a binary mask $M\in \mathbb{R}^{H\times W}$ from the BEV feature $B$. The FRPN employed in this process comprises a $3\times 3$ convolutional layer followed by a sigmoid function, rendering it exceptionally lightweight. The ground truth for this binary mask, $M^{gt}$, is derived by projecting the foreground objects onto the BEV plane.
In this paper, we use a combination of Dice loss~\cite{milletari2016v} and cross-entropy loss to supervise the FRPN.

During the inference phase, with BEV feature $B$ from F-VTM as input and the predicted binary mask $M$, we filter out unnecessary BEV grids with a mask logit lower than threshold $t_f$. Thus we obtain a  set of discrete BEV grids $\{Q_{x,y}|M[(x,y)]\!>\!t_f\}$, where $(x,y)$ is the location of each foreground BEV grid. Each BEV grid $Q_{x,y}$ can be seen as a BEV query that requires further refinement. To maintain feature consistency in the foreground area, we have selected BEV grids that contain both blank and non-blank grids.

\begin{figure}[t]
\begin{center}
\includegraphics[width=\linewidth]{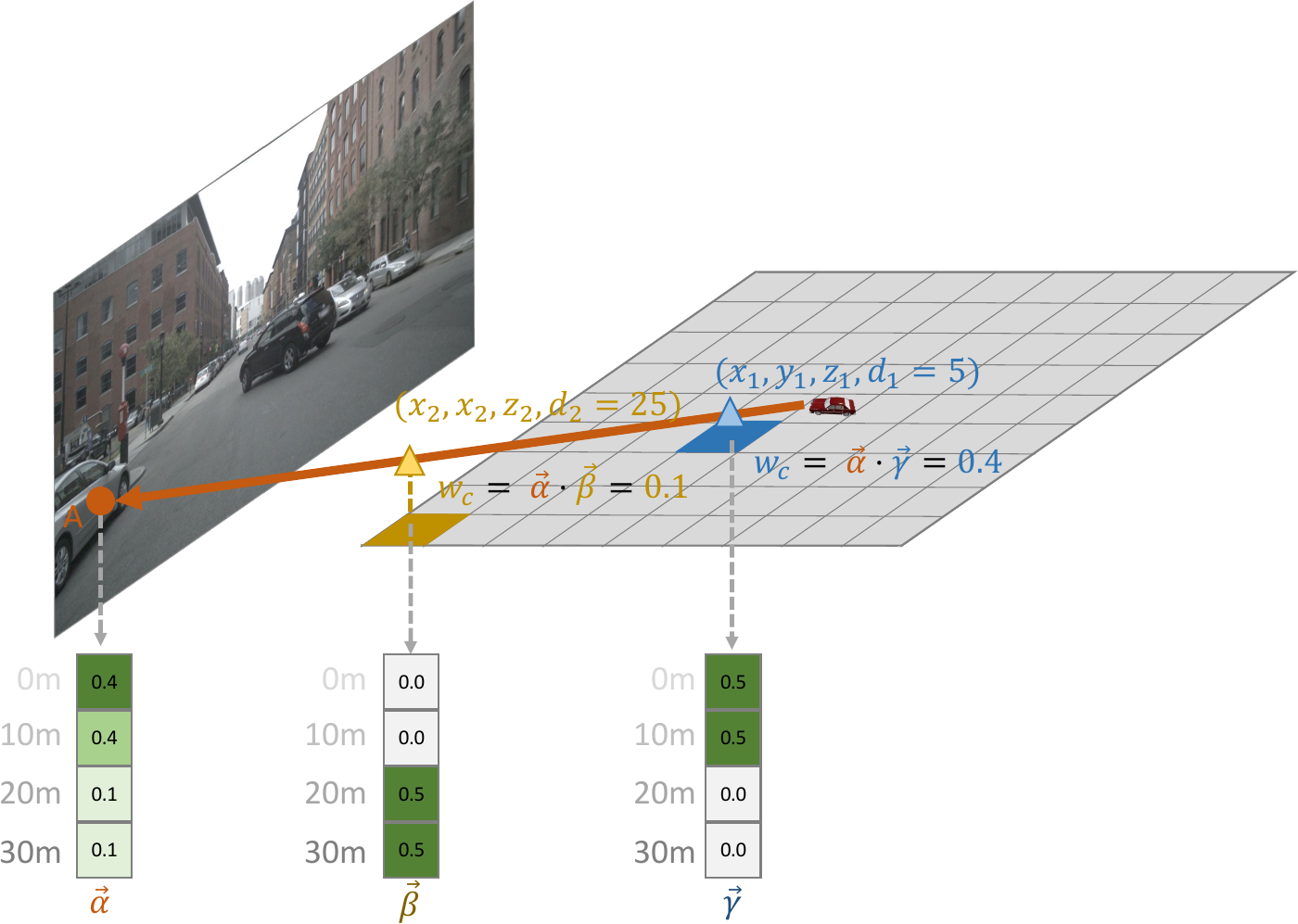}
\end{center}
   \vspace{-0.15in}
   \caption{Depth-aware backward projection uses depth consistency to distinguish features on projection rays. For instance, points $(x_1,y_1,z_1)$ and $(x_2,y_2,z_2)$ are located on the same ray, and have the same projection point $A$ on the image. The depth values of the two points are $d_1=5$m and $d_2=25$m, respectively. We then convert $d_1$ and $d_2$ to depth distribution $\Vec{\beta}$ and $\Vec{\gamma}$. Assuming the predicted depth distribution of $A$ is $\Vec{\alpha}$, the depth consistency can be computed as $\Vec{\alpha}\cdot \Vec{\gamma}=0.4$ and $\Vec{\alpha}\cdot \Vec{\beta}=0.1$. The closer point $(x_1,y_1,z_1)$ thus owns a higher feature weight with a higher consistency.}
   \label{fig:depth_aware_3d_to_2d}
   %\vspace{-0.1in}
   \end{figure}

\subsection{Depth-Aware Backward Projection}
The depth-aware backward projection module serves a dual purpose. Firstly, it effectively fills the BEV with arbitrary resolution and can choose only to generate BEV features of specified regions, thereby compensating for the sparse features generated by forward features.  Secondly, when combined with a forward projection method, they provide a more comprehensive BEV representation. In this section, we first introduce the depth consistency used to improve the quality of backward projection in \ref{sec:depth_consistecy} and then introduce our detailed implementation in \ref{sec:imple}

\subsubsection{Depth Consistency}\label{sec:depth_consistecy}
The fundamental concept of backward projection involves projecting a 3D point $(x, y, z)$ onto a 2D image point $(u, v)$, based on the camera projection matrix $P\in \mathbb{R}^{3\times 4}$. This process can be expressed mathematically as:
\vspace{-1mm}
\begin{equation}
\label{3to2_projection}
d\cdot\begin{bmatrix} u &v & 1 \end{bmatrix}^T= P \cdot \begin{bmatrix} x &y &z & 1 \end{bmatrix}^T,
\vspace{-1mm}
\end{equation}

where $d$ represents the depth of the 3D point $(x, y, z)$ on the image. Notably, for any 3D point $(\lambda x, \lambda y, \lambda z)$, where $\lambda\in \mathbb{R}^{+}$, they share the same projected point $(u, v)$ on the 2D image. Consequently, these 3D points $(\lambda x, \lambda y, \lambda z)$ exhibit similar image features, as shown in Figure~\ref{fig:3d_to_2d_vis}.

Forward projection alleviates this problem by predicting different weights for different depths. Specifically, for each point $(u, v)$, it predict a weight $w_i$ for each discrete depth $(d_0 + i\Delta)$, and $i\! \in \! \{0, 1,\cdots, |D|\}$, $D$ is a set of discrete depths, $d_0$ is the initial depth and $\Delta$ is the depth interval. Thus, while considering two discrete depth $(d_0 + i\Delta)$ and  $(d_0 + j\Delta)$ on point $(u, v)$, it falls onto the 3D points $(x_i, y_i, z_i)$ and $(x_j, y_j, z_j)$ based on Equation~\ref{3to2_projection}. The forward projection method leverages predicted depth weights $w_i$ and $w_j$ to generate distinguishing features.

To incorporate depth in backward projection and enhance the projection quality, this paper introduces depth consistency $w_c$, as shown in Figure~\ref{fig:depth_aware_3d_to_2d}. Equation~\ref{3to2_projection} shows that a 3D point $(x, y, z)$ has a corresponding depth $d\in \mathbb{R}^{+}$ on the projected point $(u, v)$. Since a discrete depth distribution vector $[w_0, w_1,\cdots, w_{|D|}]$ on point $(u,v)$ is already available. The depth consistency $w_c$ of the depth value $d$ and this depth distribution vector can be computed by converting $d$ into  depth distribution vector $[w_0',\cdots,w_i', w_{i+1}', \cdots  w_{|D|}']$, where only $w_i'$ and $w_{i+1}'$ are non zero, and $(d_0 + i\Delta) \le d \le (d_0 + (i+1)\Delta)$. The depth consistency $w_c$ can be computed as:
\vspace{-2mm}
\begin{equation}
\begin{split}
w_c &= [w_0, w_1,\cdots, w_{|D|}]\cdot [w_0', w_1',\cdots, w_{|D|}']\\
&= w_iw_i' + w_{i+1}w_{i+1}',
\end{split}
\end{equation}
where $w_i' \!= \!1\!-\!\frac{d-d_0-i\Delta}{\Delta}$ and $w_{i+1}' \!= \!1\!-\! w_i'$. 
The depth consistency introduced in this paper serves a similar role as the depth weight in forward projection. It is worth mentioning that we obtain the depth distribution of point $(u, v)$ via bilinear projection. 

Forward projection employs discrete depth values to generate corresponding discrete 3D projection points in 3D space. While sacrificing continuity in depth, the accuracy of BEV features by forward projection is also affected. 
For our depth-aware backward projection, we guarantee the ability to densely fill 3D space at arbitrary resolutions, while leveraging depth consistency to guarantee projection quality.

\subsubsection{Implementation}\label{sec:imple}
Depth consistency is a general mechanism  that can be plugged into any existing backward projection method. In this paper, our depth-aware backward projection is based on the spatial cross-attention in BEVFormer~\cite{li2022BEVFormer}. The projection process of the original Spatial Cross-Attention (SCA) of BEVFormer can be formulated as:
\vspace{-2mm}
\begin{equation}\label{sca}
    \text{SCA}(Q_{x,y}, F)\!= \!\sum_{i=1}^{{N_\text{c}}} \sum_{j=1}^{{N_\text{ref}}}
\mathcal{F}_d(Q_{x,y}, \mathcal{P}_i(x, y, z_j), F_i),
\end{equation}
where $Q_{x,y}$ is one BEV query that located at $(x, y)$ and $F$ are multi-view features. For each point $(x,y)$ on the BEV plane, BEVFormer will lift this point up to $N_{\text{ref}}$ 3D points with different heights $z_i$. Projection function will get the projection point $(u_i, v_i)$ on $i$-th image based on Equation~\ref{3to2_projection}. $\mathcal{F}_{d}$ is the deformable attention 
function~\cite{zhu2020deformable} that using query $Q_{x,y}$ to sample features of the projection point
$\mathcal{P}_i(x, y, z_j)$ on image feature $F_i$.

By using the depth consistency of this paper, we can directly evolve SCA into Depth-Aware SCA (SCA$_{da}$) by:
\vspace{-2mm}
\begin{equation}\label{sca}
    \text{SCA}_{da}(Q_{x,y}, F)\!= \!\sum_{i=1}^{{N_\text{c}}} \sum_{j=1}^{{N_\text{ref}}}
\mathcal{F}_{d}(Q_{x,y}, \mathcal{P}_i(x, y, z_j), F_i)\cdot\boldsymbol{w_c^{ij}},
\end{equation}
where $w_c^{ij}$ is the depth consistency between 3D point $(x, y, z_j)$ and 2D point $(u_i, v_i)$.
Compared to the original SCA, our proposed SCA$_{da}$ is capable of generating more discriminative BEV features along the longitudinal direction. Due to the high efficiency of our depth-aware SCA, we only use back projection once instead of stacking 6 layers used by the original BEVFormer~\cite{li2022BEVFormer}.

\section{Experiments}
\subsection{Experimental Setup}
\noindent\textbf{Dataset and Metrics.}
The nuScenes dataset~\cite{caesar2020nuscenes} is a large-scale autonomous driving dataset, which contains 1000 diverse scenarios.
Key samples of each scenario are annotated at 2Hz, and each sample includes RGB images from 6 cameras. 
NuScenes provides a series of metrics to measure the quality of 3D detection, including nuScenes Detection Score (NDS), mean Average Precision (mAP),  mean Average Translation Error (mATE), mean Average Scale Error (mASE), mean Average Orientation Error
(mAOE), mean Average Velocity Error (mAVE) and mean Average Attribute Error (mAAE).
NDS is a composite metric and defines as  ${\rm NDS}\!=\!\frac{1}{10}[5{\rm mAP}\!+\!\sum_{{\rm mTP}\in \mathbb{TP}} (1\!-\!{\rm min}(1,{\rm mTP}))]$.

\vspace{2mm}
\noindent\textbf{Implementation Details.}
By adhering to common practices~\cite{huang2021bevdet,huang2022bevdet4d,li2022bevdepth}, we default to using ResNet-50~\cite{he2016deep} and an image size of $256\!\times\! 704$. During training, we adopt the CBGS strategy~\cite{zhu2019class} and apply data augmentations at both the image and BEV levels, which include random scaling, flipping, and rotation as per BEVDet~\cite{huang2021bevdet}.
By default,  our model is trained for 20 epochs using a batch size of 64 and the AdamW~\cite{loshchilov2017decoupled} optimizer with a learning rate of 2e-4.
While training FB-BEV with V2-99 backbone for \textit{test} set, we train the model with 30 epochs without CBGS.
For training the depth net with temporal information, we use the camera-aware Depth Net in BEVDepth~\cite{li2022bevdepth} with a total of 118 depth categories ($|D|$), $d_0\!=\!1$ and $\Delta\!=\!0.5$.
Incorporating depth-aware spatial cross-attention, we sample the predefined heights uniformly from [-5m, 3m], use 8 attention heads, and set $N_{\text{ref}}\!=\!4$. The spatial shape of BEV grids is, by default, $128\!\times\! 128$ with a channel dimension of 256. The threshold for the foreground mask is set to $t_f\!=\!0.4$.
When introducing temporal information, we stack the BEV features of two adjacent keyframes, as done in BEVDet4D~\cite{huang2022bevdet4d} for \textit{val} set, and 9 previous keyframes for \textit{test} set.

\subsection{Baselines}
To assess the efficacy of our novel approach, FB-BEV, we conduct comparisons with two types of baselines that solely rely on forward and backward projection techniques, respectively. Notably, for these baselines, we maintain consistency with FB-BEV in terms of backbone, detection head, and training strategy, with the exception of the view transformation module.  We reduce the number of channels and layers of FB-BEV to match the computational cost.

\begin{table*}[t]
\centering
\caption{Comparison on the nuScenes \texttt{val} set. 
*: Baseline methods for a fair comparison.
While using depth supervision in BEVFormer,  we set training depth estimation as an auxiliary task. $\dag$: Since our channel dimension from the backbone is 256, which is half of BEVDet4D*/BEVDepth*. Thus our model is much lighter than BEVDet4D* and BEVDepth* while using the camera-aware depth net. 
%$\ddag$: R101-DCN\cite{he2016deep,dai2017deformable} initialized from a FCOS3D backbone. 
}
\vspace{-1mm}
\label{tab:main_val_set}
\renewcommand\arraystretch{0.7}

\resizebox{\textwidth}{!}{
\begin{tabular}{l|c|c|c|c|cc|c@{\hspace{1.0\tabcolsep}}c@{\hspace{1.0\tabcolsep}}c@{\hspace{1.0\tabcolsep}}c@{\hspace{1.0\tabcolsep}}c| c |c} 
% \begin{tabular}{ p{25mm}<{\centering}| p{10mm}<{\centering} | p{12mm}<{\centering}  p{10mm}<{\centering}  p{10mm}<{\centering} p{10mm}<{\centering} p{10mm}<{\centering} p{12mm}<{\centering} | p{10mm}<{\centering}}

\toprule
{Methods} & {Backbone} & {Image Size} & {Temporal}  & {Depth Sup.}& {mAP}$\uparrow$  &{NDS}$\uparrow$  & {mATE}$\downarrow$ & {mASE}$\downarrow$   &{mAOE}$\downarrow$   &{mAVE}$\downarrow$   &{mAAE}$\downarrow$  & Param. & Flops \\
\midrule
% R50 256 x 704
PETR~\cite{liu2022petr} & R50 & 384 $\!\times\!$ 1056                                      &\xmark & \xmark & 0.313 & 0.381 & 0.768 & 0.278 & 0.564 & 0.923 & {0.225} &-&-\\ % PETR Table  1
BEVDet~\cite{huang2021bevdet} & R50 & 256 $\!\times\!$ 704                                     &\xmark & \xmark & 0.298 & 0.379 & 0.725 & 0.279 & 0.589 & 0.860 & 0.245 &-\\ % BEVDet Table 7
BEVDet*& R50 & 256 $\!\times\!$ 704                                     &\xmark & \xmark & 0.307 & 0.382 & 0.722 & 0.278 & 0.606 & 0.876 & 0.235 & 55.7 & 184\\ % BEVDet Table 7
BEVFormer* & R50 & 256 $\!\times\!$ 704                                    & \xmark &\xmark & 0.297 & 0.379 & 0.739 &0.281&0.601&0.833&0.242  &59.7 & 216  \\ % BEVDet Repo, evaled mysel
\rowcolor[gray]{.9} 
FB-BEV (ours) & R50 & 256 $\!\times\!$ 704                                    & \xmark &\xmark & {0.312} & {0.406} & {0.702} & {0.275} & {0.518} & {0.777} & 0.227 & 58.4 & 192\\ % BEVDet Repo, evaled myself
\midrule
BEVDet4D~\cite{huang2022bevdet4d} & R50 & 256 $\!\times\!$ 704                                    & \cmark &\xmark & 0.322 & 0.457 & 0.703 & 0.278 & 0.495 & 0.354 & 0.206  &-&- \\ % BEVDet Repo, evaled myself
BEVDet4D*  & R50 & 256 $\!\times\!$ 704                                    & \cmark&\xmark & 0.344 & 0.466 & 0.670 & 0.273 & 0.523 & 0.400 & 0.194& 83.4$^\dag$ & 296 \\ % BEVDet Repo, evaled myself
BEVFormer-T*  & R50 & 256 $\!\times\!$ 704                                    & \cmark&\xmark 
&0.330 &0.459 & 0.686&0.272&0.482&0.417&0.201& 66.9 & 249 \\ % BEVDet Repo, evaled myself
\rowcolor[gray]{.9} 
FB-BEV (ours)  & R50 & 256 $\!\times\!$ 704                                    & \cmark&\xmark & 0.350&0.479 &0.642&0.275&0.459&0.391&0.193 &65.7 &225\\ % BEVDet Repo, evaled myself

\midrule

STS~\cite{wang2022sts} & R50      & 256 $\!\times\!$ 704  & \cmark& \cmark & 0.377 & 0.489 & 0.601 & 0.275 & 0.450 & 0.446 & 0.212 &-&-\\ % STS Table 1
% BEVStereo & R50 & 256 $\!\times\!$ 704  & \cmark& \cmark & 0.372 & 0.500 & 0.598 & 0.270 & 0.438 & 0.367 & 0.190 &-\\ % BEVStereo Github
BEVDepth~\cite{li2022bevdepth} & R50 & 256 $\!\times\!$ 704  & \cmark& \cmark & 0.351 & 0.475 & 0.639 & {0.267} & 0.479 & 0.428 & 0.198&-&- \\ % BEVDepth Table 4
BEVDepth* & R50 & 256 $\!\times\!$ 704  & \cmark& \cmark & 0.370 & 0.484 & 0.611 & 0.271 & 0.493 & 0.423 & 0.211 &83.4$^\dag$ & 292\\ % BEVDepth Table 4
BEVFormer-T*& R50 & 256 $\!\times\!$ 704  & \cmark& \cmark &0.343 &0.461 & 0.680&0.274&0.519&0.426&0.204&66.9 &249 \\ % BEVDepth Table 4
\rowcolor[gray]{.9} 
FB-BEV (ours)& R50 & 256 $\!\times\!$ 704  & \cmark& \cmark & 0.378 & 0.498&0.620&0.273&0.444& 0.374&0.200 &65.7 & 225\\ % BEVDepth Table 4

\bottomrule
\end{tabular}
}
\vspace{-1mm}
\end{table*}
\begin{table*}[t]
\centering
\caption{Comparison on the nuScenes \texttt{test} set. Extra data is depth pertaining. V2-99 ~\cite{lee2019energy,park2021pseudo} uses extra data for depth training. Swin-B~\cite{liu2021swin} and ConvNeXt-B~\cite{liu2022convnet} are trained with ImageNet-22K~\cite{deng2009imagenet}.}
\label{tab:main_test_set}
\vspace{-1mm}
\renewcommand\arraystretch{0.7}
\resizebox{\textwidth}{!}{
\begin{tabular}{l|c|c|c|c|c|c|ccccc} 
% \begin{tabular}{ p{25mm}<{\centering}| p{10mm}<{\centering} | p{12mm}<{\centering}  p{10mm}<{\centering}  p{10mm}<{\centering} p{10mm}<{\centering} p{10mm}<{\centering} p{12mm}<{\centering} | p{10mm}<{\centering}}

\toprule
{Methods} & {Backbone} & {Image Size} &   {Test-Time Aug} & {mAP}$\uparrow$  &{NDS}$\uparrow$  & {mATE}$\downarrow$ & {mASE}$\downarrow$   &{mAOE}$\downarrow$   &{mAVE}$\downarrow$   &{mAAE}$\downarrow$  \\
\midrule
FCOS3D~\cite{wang2021fcos3d}         & R101-D   & 900$\!\times\!$1600  & \cmark & 0.358 & 0.428 & 0.690 & 0.249 & 0.452 & 1.434 & 0.124 \\
DETR3D~\cite{wang2022detr3d}        & V2-99     & 900$\!\times\!$1600 & \cmark & 0.412 & 0.479 & 0.641 & 0.255 & 0.394 & 0.845 & 0.133 \\
UVTR~\cite{li2022uvtr}           & V2-99      & 900$\!\times\!$1600 & \xmark & 0.472 & 0.551 & 0.577 & 0.253 & 0.391 & 0.508 & 0.123 \\
BEVFormer~\cite{li2022BEVFormer}      & V2-99      & 900$\!\times\!$1600  & \xmark & 0.481 & 0.569 & 0.582 & 0.256 & 0.375 & 0.378 & 0.126 \\
BEVDet4D~\cite{huang2022bevdet4d}       & Swin-B     & 900$\!\times\!$1600  & \cmark & 0.451 & 0.569 & 0.511 & \textbf{0.241} & 0.386 & 0.301 & {0.121} \\
PolarFormer~\cite{jiang2022polarformer}    & V2-99      & 900$\!\times\!$1600 & \xmark & 0.493 & 0.572 & 0.556 & 0.256 & 0.364 & 0.439 & 0.127 \\
PETRv2~\cite{liu2022petrv2}         & V2-99  & 640$\!\times\!$1600  & \xmark & 0.490 & 0.582 & 0.561 & 0.243 & 0.361 & 0.343 & \textbf{0.120} \\
BEVStereo~\cite{li2022bevstereo}      & V2-99      & 640$\!\times\!$1600  & \xmark & 0.525 & 0.610 & \textbf{0.431} & 0.246 & \textbf{0.358} & 0.357 & 0.138 \\
BEVDepth~\cite{li2022bevdepth}       & V2-99 & 640$\!\times\!$1600 & \xmark 

& 0.503
&0.600
& 0.445
& 0.245
& 0.378
&0.320
&0.126
 \\
%BEVDepth~\cite{li2022bevdepth}       & ConvNeXt-B~\cite{liu2022convnet} & 640$\!\times\!$1600 & \xmark & \xmark & 0.520 & 0.609 & 0.445 & 0.243 & {0.352} & 0.347 & 0.127 \\
SOLOFusion~\cite{park2022time} & ConvNeXt-B & 640$\!\times\!$1600 &\xmark & \textbf{0.540} &0.619 &0.453 &0.257& 0.376& 0.276& 0.148\\

\rowcolor[gray]{.9} 

% running
FB-BEV (ours)     & V2-99 & 640$\!\times\!$1600 & \xmark &0.537 &\textbf{0.624} &0.439 & 0.250 & \textbf{0.358} & \textbf{0.270} & 0.128      \\

% "mAP": 0.534116746149049, "mATE": 0.45328743248713677, "mASE": 0.25589972106417147, "mAOE": 0.3731649874352068, "mAVE": 0.28226529667475686, "mAAE": 0.1307984811895046, "NDS": 0.6175167811894469}}

\bottomrule
\end{tabular}
}
\vspace{-2mm}
\end{table*}
\begin{table*}[t]
\centering

\caption{Effect of depth in backward projection.}
\vspace{-1mm}
\renewcommand\arraystretch{0.7}

\resizebox{\textwidth}{!}{
\begin{tabular}{l|c|c|c|c|c|c|c@{\hspace{1.0\tabcolsep}}c@{\hspace{1.0\tabcolsep}}c@{\hspace{1.0\tabcolsep}}c@{\hspace{1.0\tabcolsep}}c} 
% \begin{tabular}{ p{25mm}<{\centering}| p{10mm}<{\centering} | p{12mm}<{\centering}  p{10mm}<{\centering}  p{10mm}<{\centering} p{10mm}<{\centering} p{10mm}<{\centering} p{12mm}<{\centering} | p{10mm}<{\centering}}

\toprule

\textbf{Methods} & \textbf{Backbone} & \textbf{Image Size} & \textbf{Temporal} & \textbf{Depth Sup.} & \textbf{mAP}$\uparrow$  &\textbf{NDS}$\uparrow$  & 
\textbf{mATE}$\downarrow$ & \textbf{mASE}$\downarrow$   &\textbf{mAOE}$\downarrow$   &\textbf{mAVE}$\downarrow$   &\textbf{mAAE}$\downarrow$  \\

% \midrule
% BEVDet & R50 & 256$\!\times\!$704 &\xmark & \xmark & 0.307 & 0.382 & 0.722 &0.278&0.606&0.876&0.235\\
% w/ uniform depth & R50 & 256$\!\times\!$704 &\xmark & \xmark & 0.297&0.360&0.758 &0.281&0.633&0.924&0.288\\

\midrule
BEVFormer & R50 & 256$\!\times\!$704 &\xmark & \xmark & 0.297 & 0.379 & 0.739 &0.281&0.601&0.833&0.242\\
w/ depth-aware & R50 & 256$\!\times\!$704 &\xmark & \xmark & 0.291&0.387&0.740 &0.282&0.548&0.806&0.225\\
\midrule
BEVFormer & R50 & 256$\!\times\!$704 &\cmark & \cmark &0.343 &0.461 & 0.680&0.274&0.519&0.426&0.204\\
w/ depth-aware & R50 & 256$\!\times\!$704 &\cmark & \cmark & 0.350&0.472&0.665 &0.281&0.499&0.390&0.194\\

\midrule
FB-BEV & R50 & 256$\!\times\!$704 &\xmark & \xmark & 0.312 & 0.406 & 0.702 & 0.275 & 0.518 & 0.777 & 0.227\\
w/o depth-aware & R50 & 256$\!\times\!$704 &\xmark & \xmark & 0.305&0.397&0.726 &0.278&0.552&0.779&0.227\\
\midrule
FB-BEV & R50 & 256$\!\times\!$704 &\cmark & \cmark  & 0.378 & 0.498&0.620&0.273&0.444& 0.374&0.200 \\
w/o depth-aware & R50 & 256$\!\times\!$704 &\cmark & \cmark & 0.367&0.489&0.629&0.273&0.458&0.382&0.196\\
\bottomrule
\end{tabular}
}
% \vspace{-3mm}
\label{tab:abl_depth_aware}
\end{table*}

\vspace{2mm}
\noindent\textbf{Forward Projection.}
For forward projection methods, we adopt BEVDet~\cite{huang2021bevdet,huang2022bevdet4d} and BEVDepth~\cite{li2022bevdepth} as our baseline.
Compared to BEVDet, BEVDepth uses point clouds to generate the ground truth of depth and train the depth net with the ground truth of depth.

\vspace{2mm}
\noindent\textbf{Backward Projection.}
For backward projection methods, we choose BEVFormer~\cite{li2022BEVFormer} as the baseline. Considering the difference in implementation details, we ported the view transformation module of BEVFormer to BEVDet for a fair comparison. It is worth mentioning that we discard the temporal self-attention module in BEVFormer. In this paper, we note the BEVFormer that with temporal information as BEVFormer-T.

\subsection{Benchmark Results}
Table~\ref{tab:main_val_set} shows the 3D detection results on the nuScenes \texttt{val} set for our proposed FB-BEV method, as well as the two baseline methods BEVDet~\cite{huang2021bevdet} and BEVFormer~\cite{li2022BEVFormer}, and other previous state-of-the-art 3D detection methods. 
Without using temporal information or depth supervision, our method outperforms BEVDet and BEVFormer by a significant margin of 2.4\% NDS and 2.7\% NDS.
When introducing temporal information by stacking historical BEV features, our proposed FB-BEV still outperforms BEVDet and BEVFormer by 1.3 points. With depth supervision, our method achieves a lead of more than 1.5 points over BEVDepth. However, as the previous backward projection cannot use depth information, BEVFormer-T only brings a marginal improvement of 0.2\% NDS when only using depth supervision as an auxiliary task. This confirms the limitations of existing backward projection methods.
Despite achieving higher performance, our method still maintains a comparable or even lower computational cost than our baselines. 
As shown in Table~\ref{tab:main_test_set}, our model obtains a new state-of-the-art 62.4\% NDS and outperforms previous SOLOFusion~\cite{park2022time} with a clear margin of 0.5 points.

\subsection{Ablation Studies}
\noindent\textbf{Depth-aware Backward Projection.}
In Table~\ref{tab:abl_depth_aware}, we compare the results of adopting depth-aware backward projection in FB-BEV and BEVFormer. BEVFormer obtains an improvement of 0.9\% NDS with depth-aware projection. When using depth supervision as an auxiliary task, BEVFormer-T achieves a larger gain of 1.1\% NDS. Without depth-aware backward projection in FB-BEV, the performance drops by about 0.9\% NDS.
In the past, only forward projection methods could benefit from more accurate depth prediction. With depth consistency, backward methods can also improve performance by leveraging accurate depth prediction.

\begin{figure}[htb]
	\begin{center}
	\includegraphics[width=\linewidth]{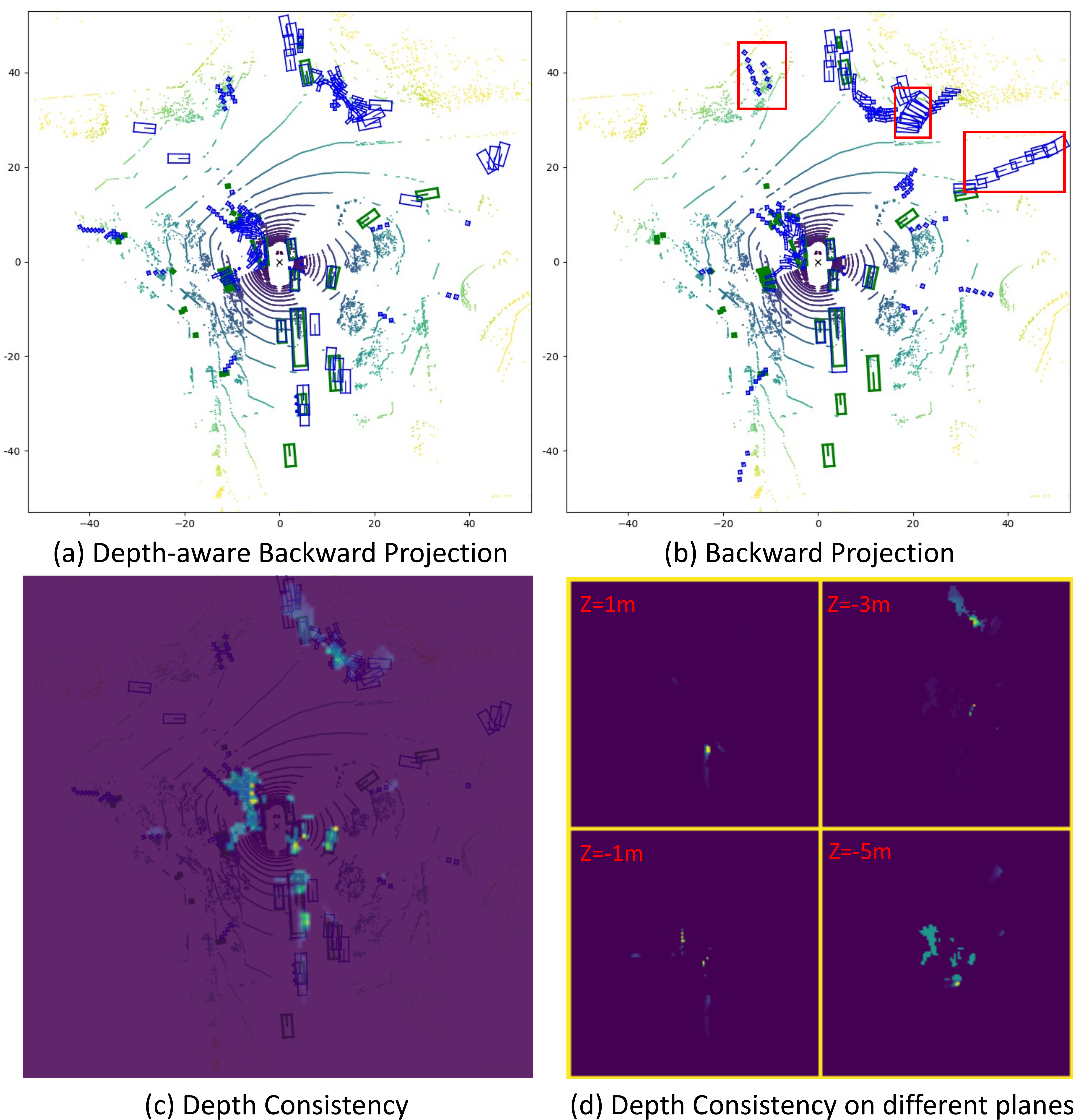}
	\end{center}
	\vspace{-0.1in}
	\caption{(a)/(b) Comparison of FB-BEV back projection with and without depth consistency, respectively. The red boxes in (b) indicate erroneous results produced by the model along the longitudinal direction due to depth ambiguities. (c) shows the depth consistency map on the BEV plane, where each value is the sum of depth consistency at different heights. In (d), we observe the depth consistency map at different heights.
	}
	\label{fig:abl_depth_aware}
	\vspace{-1mm}
\end{figure}

Figure~\ref{fig:abl_depth_aware} presents visual results of FB-BEV with and without depth-aware backward projection. When the depth-aware projection is not employed, the model tends to produce incorrect results along the longitudinal direction due to depth ambiguities, as seen in the yellow boxes in Figure~\ref{fig:abl_depth_aware}~(b). In addition, Figure~\ref{fig:abl_depth_aware}~(c) and (d) show the depth consistency on the BEV plane for FB-BEV with and without depth-aware projection. Foreground grids exhibit higher depth consistency, which prevents background regions from erroneously identifying false foreground features. Moreover, Figure~\ref{fig:abl_depth_aware}~(d) shows that the depth consistency varies with height for the same location $(x,y)$ on the BEV plane. Prior backward projection methods aggregated features at all heights, resulting in feature interference. However, with our proposed depth-aware backward projection, the model selectively aggregates features based on depth consistency at different heights. These visualizations provide compelling evidence for the effectiveness of our method.

\begin{figure}[htb]
	\begin{center}
	\includegraphics[width=\linewidth]{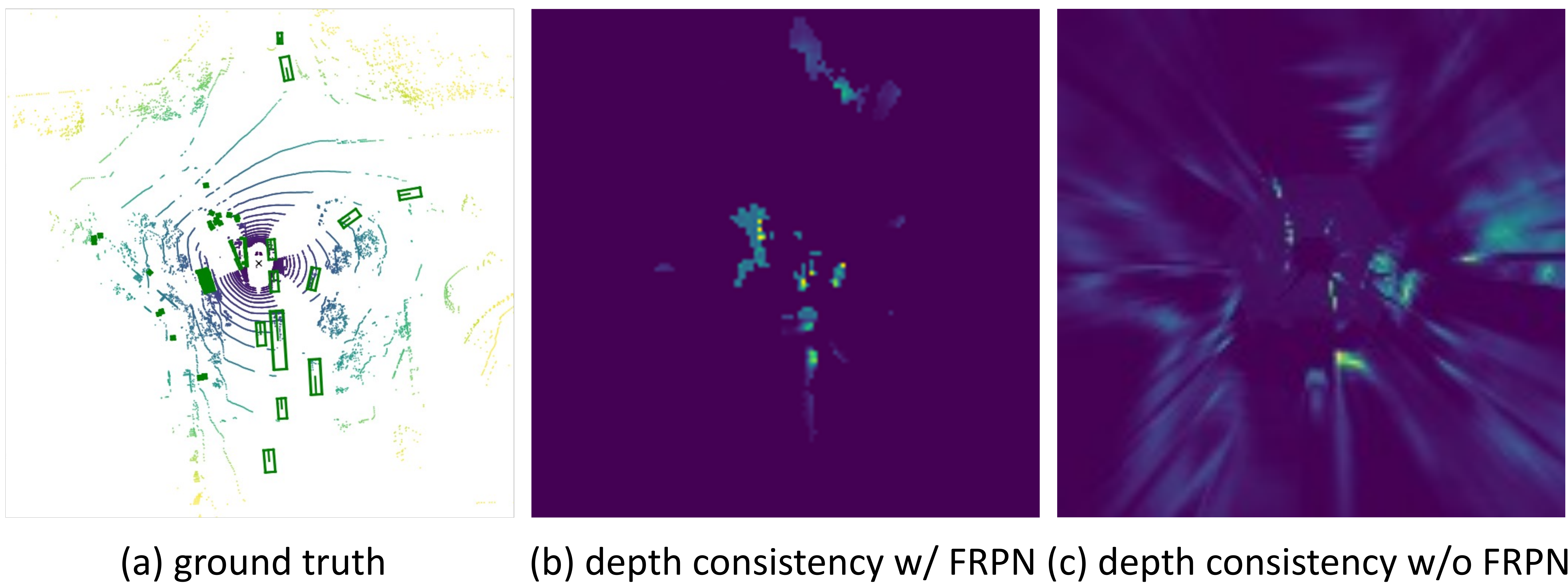}
	\end{center}
	\vspace{-0.1in}
	\caption{We compare the depth consistency map of FB-BEV with FRPN or not in (b) and (c).  In the absence of FRPN, depth-aware backward projection may still attend to certain background regions due to the presence of inaccurate depth prediction. On the other hand, FRPN utilizes the foreground mask to enable the model to concentrate solely on the foreground objects.}
	\label{fig:frpn}
	\vspace{-1mm}
\end{figure}

\vspace{2mm}
\noindent\textbf{Effect of FRPN.}
We employ FRPN to selectively optimize the foreground grids in BEV feature $B$ through B-VTM. To study its effectiveness, we conduct an experiment where we exclude FRPN and instead feed all BEV features into B-VTM. Results in Table~\ref{tab:abl_sparse} demonstrate that FRPN not only improves the inference efficiency but also improves the detection accuracy. In Figure~\ref{fig:frpn}, we present the depth consistency map of FB-BEV with and without FRPN. Without using FRPN, depth-aware backward projectionmay focus on some background regions due to imprecise depth predictions. On the other hand, using the foreground mask provided by FRPN, the model can selectively concentrate only on the foreground objects, thus avoiding interference from the background regions.

\begin{table}[t]
\centering
\caption{Effect of FRPN. With FRPN, FB-BEV obtains higher accuracy and faster inference. The latency of VTM is smaller due to only refining the foreground BEV grids.}
\label{tab:abl_sparse}
\vspace{-1mm}
\renewcommand\arraystretch{0.7}
\resizebox{\linewidth}{!}{
\begin{tabular}{l|c|c|c|c|c} 
% \begin{tabular}{ p{25mm}<{\centering}| p{10mm}<{\centering} | p{12mm}<{\centering}  p{10mm}<{\centering}  p{10mm}<{\centering} p{10mm}<{\centering} p{10mm}<{\centering} p{12mm}<{\centering} | p{10mm}<{\centering}}

\toprule

\textbf{Methods} & \textbf{Temporal} & \textbf{Depth Sup.}  & \textbf{mAP}$\uparrow$  &\textbf{NDS}$\uparrow$  & Latency\\
\midrule

FB-BEV  &\xmark & \xmark &0.312 & 0.406 & 2.6ms\\
w/o FRPN & \xmark & \xmark & 0.308&0.400 & 3.4ms \\
\midrule
FB-BEV &\cmark & \cmark  & 0.378 & 0.498& 2.6ms \\
w/o FRPN & \cmark & \cmark & 0.373&0.494& 3.4ms \\

\bottomrule
\end{tabular}
}
\vspace{-1mm}
\end{table}

\begin{table}[t]
\centering
\caption{Ablation on the Effect of BEV scale. All results are trained for 12 epochs with CBGS. We also show the latency of VTM during inference. }
\label{tab:abl}
\vspace{-1mm}
\renewcommand\arraystretch{0.7}
\setlength{\tabcolsep}{4mm}
\resizebox{\linewidth}{!}{
\begin{tabular}{l|c|c|c|c} 
\toprule
\textbf{Methods} & \textbf{BEV Size}   & \textbf{mAP}$\uparrow$  &\textbf{NDS}$\uparrow$  & Latency \\
\midrule
BEVDet  &128$\!\times\!$128 &0.304 & 0.370 &0.7ms \\
FB-BEV  &128$\!\times\!$128 &0.309 & 0.396 & 2.6ms \\
\midrule
BEVDet  &256$\!\times\!$256 &0.309 & 0.375& 0.8ms\\
FB-BEV  &256$\!\times\!$256 &0.322 & 0.404 &3.4ms \\
\midrule
BEVDet  &400$\!\times\!$400 &0.302 & 0.368 &1.5ms \\
FB-BEV  &400$\!\times\!$400 & 0.325 &0.406 &6.6ms \\
\bottomrule
\end{tabular}
}
\end{table}

\vspace{2mm}
\noindent\textbf{Effect of Reducing Sparsity.}
Due to the fixed discrete depth values, the forward projection method generates fixed discrete 3D projection points (projection matrix $P$ remains unchanged). As the BEV scale increases, the proportion of blank grids on BEV generated by forward projection will also increase. The rate of blank gird of BEVDet with a BEV scale $400\!\times\! 400$ and input shape $256\!\times\!704$ is 80.5\%. Thus we can observe that BEVDet performance drops on large-scale BEV. Our forward-backward projection fixes it by filling these blank grids, then obtains continuous performance gains. In addition,  we can observe that current VTMs are highly efficient and are not considered a potential bottleneck against inference efficiency.

\section{Conclusion}
We present a forward-backward projection paradigm to address the limitations of current projection schemes. Our approach addresses the issue of sparse features generated by forward projection and introduces depth into backward projection to establish a more precise projection relationship. 
This two-stage VTM strategy is suitable for higher-resolution BEV perception and has application prospects for ultra-long-distance object detection or high-resolution occupancy perception.

% \section*{Acknowledgements}
% This work is supported by the Natural Science Foundation of
% China under Grant 61672273 and Grant 61832008.

{\small
\bibliographystyle{unsrt}
\bibliography{egbib}
}

\end{document}